# AIoT-BASED SMART TRAFFIC MANAGEMENT SYSTEM

Ahmed Mahmoud Elbasha and Mohammad M. Abdellatif

Electrical Engineering Department, Faculty of Engineering, The British University in Egypt, Cairo, Egypt

*ABSTRACT*

*This paper presents a novel AI-based smart traffic management system de-signed to optimize traffic flow and reduce congestion in urban environments. By analysing live footage from existing CCTV cameras, this approach eliminates the need for additional hardware, thereby minimizing both deployment costs and ongoing maintenance expenses. The AI model processes live video feeds to accurately count vehicles and assess traffic density, allowing for adaptive signal control that prioritizes directions with higher traffic volumes. This real-time adaptability ensures smoother traffic flow, reduces congestion, and minimizes waiting times for drivers. Additionally, the proposed system is simulated using PyGame to evaluate its performance under various traffic conditions. The simulation results demonstrate that the AI-based system out-performs traditional static traffic light systems by 34%, leading to significant improvements in traffic flow efficiency. The use of AI to optimize traffic signals can play a crucial role in addressing urban traffic challenges, offering a cost-effective, scalable, and efficient solution for modern cities. This innovative system represents a key advancement in the field of smart city infra-structure and intelligent transportation systems.*

## 1. INTRODUCTION

The increase in traffic congestion not only challenges the effectiveness of transportation systems but it has significant negative impacts on human health. Such as, the increase in travel costs, high level of anxiety among travellers, and increased air pollution [1]. Understanding the issues and difficulties has been considered by researchers and authorities globally to discover a solution to im-prove traffic congestion [2].

New actions were taken for development and implementation using enhanced technologies to improve traffic management, improve public transportation systems, the upgrade of sustainable modes of travel, and integrating smart traffic management systems that can adjust to changing circumstances, ensuring a road network that is safer and more effective [3].

Smart Traffic Management Systems represent an effective solution for traffic congestion and have advanced technologies such as harnessing the power of the Internet of Things (IoT) and Artificial Intelligence (AI) which have been increasing worldwide.

The integration of an IoT-based Smart Traffic Management System including AI algorithms has been a transformative approach to address the complicated challenges proposed by growing urban traffic. The implementation of interconnected sensors and devices along roadways would enable real-time data collection on traffic flow, vehicle movements, and environmental conditions which is conducted by an IoT-based system. Therefore, the data is given to an AI algorithm that will analyse patterns, expect congestion hotspots, and adjust traffic signals and routes. We live in an era that is characterized by fast urbanization and increasing demands on transportation networks, therefore, the integration of IoT technology into traffic management has emerged as a transformative approach to those complex issues. The power of AI will adjust the traffic flow, reduce traffic jams, and boost overall transportation efficiency.

The benefits of such advanced systems expand the congestion improvements as they would create safer road environments that will enable intelligent traffic monitoring, early detection of possible hazards, and an instant response mechanism. Furthermore, implementing IoT and AI in traffic management in smart cities is associated with increasing technologies that are improving the quality of life of the citizens and promoting sustainable urban development. The objectives of this paper are:

- Design a smart traffic control system that adapts signal timing based on current conditions.
- Improve traffic flow efficiency by optimizing signal timings and traffic patterns.

The rest of the paper is organized as follows. Section 2 gives a brief background on the topic of smart traffic management. Section 3 states the methodology followed when creating this work. Section 4 describes experimental work and lists the main results. and finally, Section 5 Concludes the paper.

## 2. BACKGROUND

The development and implementation of IoT-based Smart Traffic Management Systems have gotten significant attention in recent years [4]. Therefore, the choice of machine learning techniques is important to an efficient AI algorithm for traffic data analysis and prediction. Also, choosing the right algorithms plays a crucial role in determining the accuracy and dependability of traffic forecasts.

In [5], the authors discussed the pivotal role of wireless communication technologies, particularly in the context of 5G and 6G networks, in enabling intelligent transportation systems (ITS). It identifies three main technological pillars crucial for future ITS deployment: edge computing solutions to manage the vast data exchange required for vehicle-to-everything (V2X) communications, integrated communication and sensing (ICAS) systems that improve vehicular communication and environment mapping, and advances in cellular based side link communications for enhanced data delivery. Moreover, the paper focused on evaluating key advancements in edge computing, ICAS, and side link communications for V2X systems. The paper highlights the application of edge computing through vehicular edge computing (VEC), which manages content offloading and caching to support low-latency communications for road safety and infotainment. For ICAS, the paper discusses how joint communication, and sensing can enhance environmental mapping and vehicle localization by exploiting V2X signals. The work also investigates 5G-V2X side link communication performance in both sub-6 GHz and mmWave frequency ranges, showing that higher frequencies allow for improved accuracy but suffer from shorter coverage.

Abdul Kadar et al. [6] proposed a real-time IoT-based traffic management system for enhancing efficiency and safety. They stated that the issues of traffic congestion led to time and fuel wastage, environmental pollution, and impediments to emergency vehicle movements. They address this problem with an innovative solution in the form of a traffic management system (TMS) that uses data analytics and the Internet of Things (IoT) in real-time. Therefore, their system employs ultrasonic sensors to estimate traffic density and then collects data to analyse by a system controller, which, in turn, determines ideal traffic signal timings through a traffic management algorithm. The communication of the system works by sending the information to a cloud server via a Wi-Fi module, acknowledging the prediction of possible congestion at intersection points. Notably, emergency vehicles are prioritized by extending signal duration to facilitate their passage. To impose traffic regulations, they noted that their system can identify signal violators by giving them fines through a Traffic Wallet mobile application. The proposed system of TMS is a cost-effective solution with simplicity in installation and maintenance that helps with the challenges posed by burgeoning traffic congestion.

A comprehensive study by Jiyuan et al. [7] on deep reinforcement learning offers a promising method for solving complicated signal optimization problems at junctions, that is enabled by advanced detector technology that provides highly accurate traffic data. This study focuses on extending their model using advanced sensor technology as it explores variations of different scenarios under traffic and then analyses the connection between actual conditions and traffic demand.

Khan et al. in [8] introduced a conventional system that adjusts green light timings generally, irrespective of the actual traffic rate at junctions. Therefore, this article presents a machine learning-based, self-adaptive real-time traffic light control algorithm and image processing techniques to enhance traffic flow management at signalized junctions. The authors proposed an approach utilizing the You Only Look Once (YOLOv3) framework and a neural network to assess and manage traffic clearance at signalized intersections. The system used four deliberately positioned cameras at the top of a road junction and those cameras transmit real-time images to an embedded controller that afterward performs various operations.

Additionally, the cameras capture one side of the road with the system functionalities that include instructing the cameras to take pictures of the traffic situation, processing the input images to count the number of vehicles, determining the duration for which the green light should be active based on the vehicle count, and finally, illuminating the green light on the corresponding signal while turning the red lights on the other signals. The system optimally uses those images for accurate vehicle detection and traffic signal control. It was noted that YOLO runs significantly faster compared to alternative detecting techniques that can process a single image in an average of 1.3 seconds. To maximize the duration of the green light, factors such as road width, the number of two or four-wheel drive vehicles, and junction crossing time are considered. for effective traffic clearance. Then the system adapts to real-time traffic conditions using the input from YOLO to dynamically adjust the green light duration. The system is trained on diverse datasets, such as real traffic images being utilized to evaluate the performance giving high accuracy results.

## 3. RESEARCH METHODOLOGY

The methodology for this work involves developing a smart traffic management system using YOLOv3 [9] for vehicle detection from CCTV footage and Pygame [10] for simulation.

Pygame is a set of Python modules designed for writing video games. It provides functionality for handling multimedia elements such as graphics, sound, and input devices in a game environment. Pygame is built on top of the Simple Direct Media Layer (SDL), which gives it access to low-level hardware functionalities, making it suitable for creating both 2D and simple 3D games. Additionally, Pygame is widely used beyond game development, including applications like simulations, multimedia programs, and interactive educational tools. In this work, Pygame is used to create a custom simulation that mimics real-world traffic. It helps with system visualization and system comparison with the current static system. In the simulation, there's a four-way intersection with traffic lights at each corner. Above each light, a timer shows how much time is left before it changes from green to yellow, yellow to red, or red to green. Next to each light, the number of vehicles that have passed through the intersection is also displayed. Vehicles such as cars, motorcycles, buses, trucks, and rickshaws come from all directions. Some cars in the rightmost lane turn to cross the intersection, making the simulation feel more realistic. When a new vehicle appears, a random number decides if it will turn. There's also a timer that tracks how much time has passed since the simulation started.

The system collects real-time videos from traffic cameras and historical traffic data to train and test the YOLOv3 model for accurate vehicle identification. This real-time vehicle count data is fed to an adaptive algorithm to adjust traffic signal timings dynamically. These adaptive controls are simulated in Pygame [10] to visualize and optimize traffic flow under various conditions.

The system works by capturing an image from CCTV cameras at traffic lights to measure traffic density in real-time using image processing and object detecion. As shown in Fig. 1, the image is then analysed by a vehicle detection algorithm powered by YOLO. It identifies the number of vehicles in different categories like cars, bikes, buses, and trucks. This information helps calculate the traffic density. A signal control algorithm then uses this data, along with other factors, to decide how long the green light should stay on for each intersection. The red-light timings are adjusted automatically. To keep things fair, the green light duration is capped at both a maximum and minimum limit, ensuring no intersection is left waiting too long.

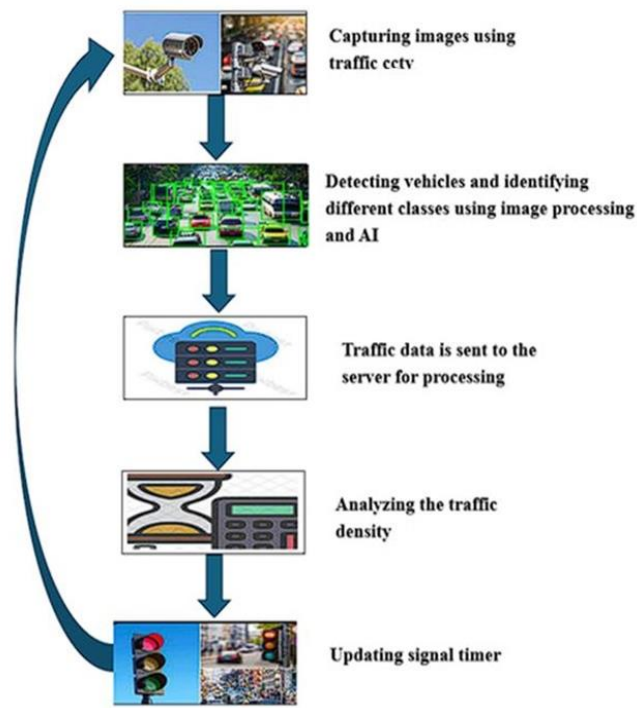

Figure 1. Flow diagram of the system.

## 4. EXPERIMENTAL WORK AND MAIN RESULTS

### 4.1. Vehicle Detection

Fig. 2 presents the test images used with our vehicle detection model. The left side includes original images, while the results after applying the model, includ-ing bounding boxes and their labels are shown on the right.

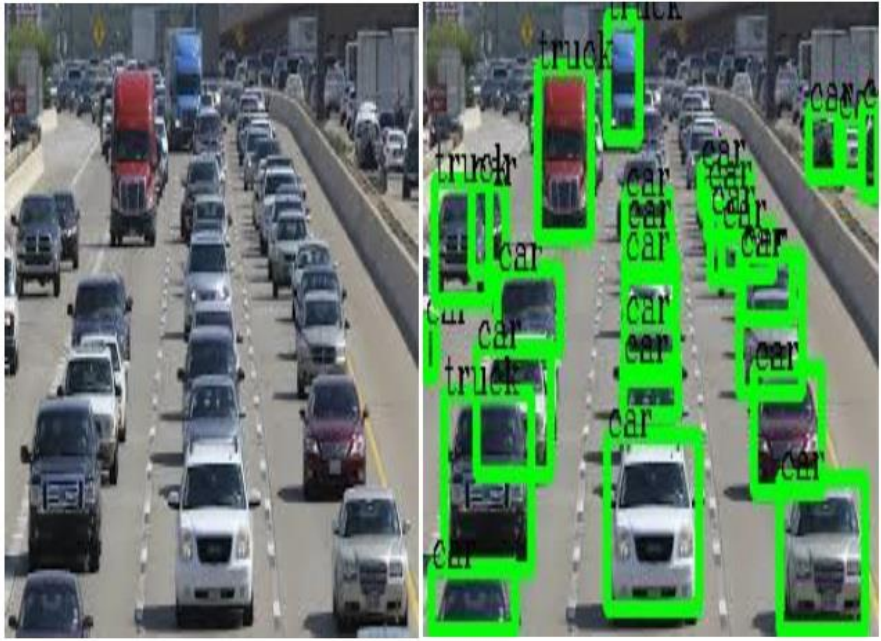

Figure 2. Vehicle Detection Results.

### 4.2. Switching Algorithm

The Signal Switching Algorithm sets the green signal timer according to traffic density returned by the vehicle detection module and updates the red signal timers of other signals accordingly. It also switches between the signals cyclically according to the timers. The algorithm was developed by considering the following factors:

- The time spent to calculate traffic density to calculate the green light duration, which helps decide when to capture the image.
- The number of lanes.
- The total number of vehicles in each category, such as cars, trucks, and motorcycles.

The green signal time is then calculated using the following formula:

$$GST = \frac{\sum_{VC}(NoOfVehicles_{VC} * Averagetime_{VC})}{NoOfLanes + 1}$$

(1)

Where,

- GST: green signal time.

- VC: Vehicle Class.
- NoOfVehicles_VC: The total count of vehicles from each class at the intersection.
- Averagetime_VC: The estimated time it takes for each class of vehicle to pass through the intersection.
- NoOfLanes: The number of lanes available at the intersection for each vehicle class.

**4.3. Simulation Model**

A simulation was created using Pygame to represent real-life traffic. It helps visualize the system and compare it with the traditional static system. The setup includes a 4-way intersection with four traffic signals. A timer is placed next to each traffic light, the timer shows the remaining time before switching to the next colour. Next to each signal, whenever a vehicle crosses the intersection, a count is increased and displayed. Fig. 3 shows a screenshot of the initial setup in the simulator of the intersection.

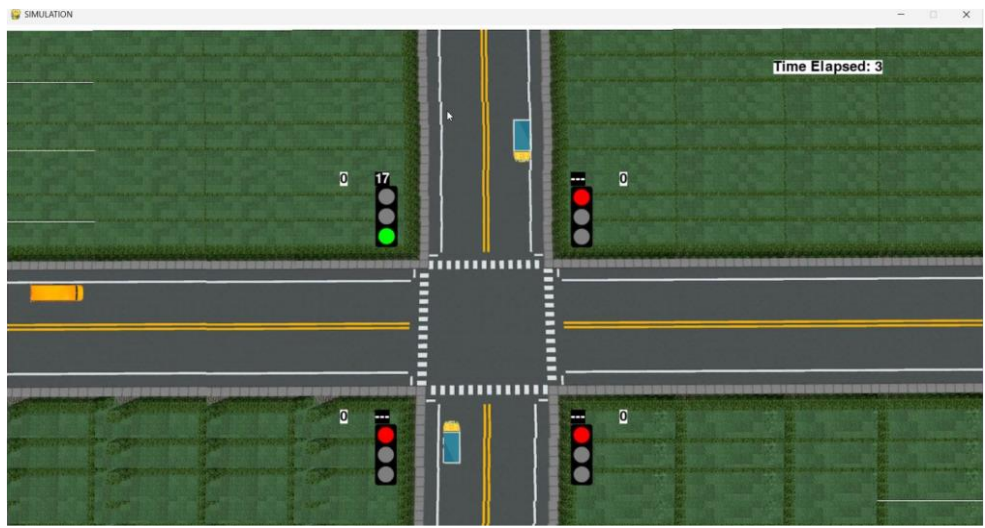

Figure 3. Initial setup of the intersection in the simulation.

The simulation begins with traffic lights displaying red and green signals. The green signal's countdown starts from a default value of 20 seconds, while the red signal's countdown is initially blank. Once the red signal countdown reaches 10 seconds, the countdown becomes visible. Next to each traffic light, a number is displayed that shows the number of vehicles who has crossed the intersection so far, starting at 0 for both signals. Additionally, the elapsed time from the start of the simulation is shown in the top right corner.

Vehicles of all types can enter the intersection from all directions. To make the simulation more realistic, some vehicles in the rightmost lane turn to cross the intersection. The decision to turn is decided at random per vehicle. The simulation also includes a timer showing the elapsed time from the beginning of the simulation. Fig. 4 shows a screenshot of the final output of the simulation.

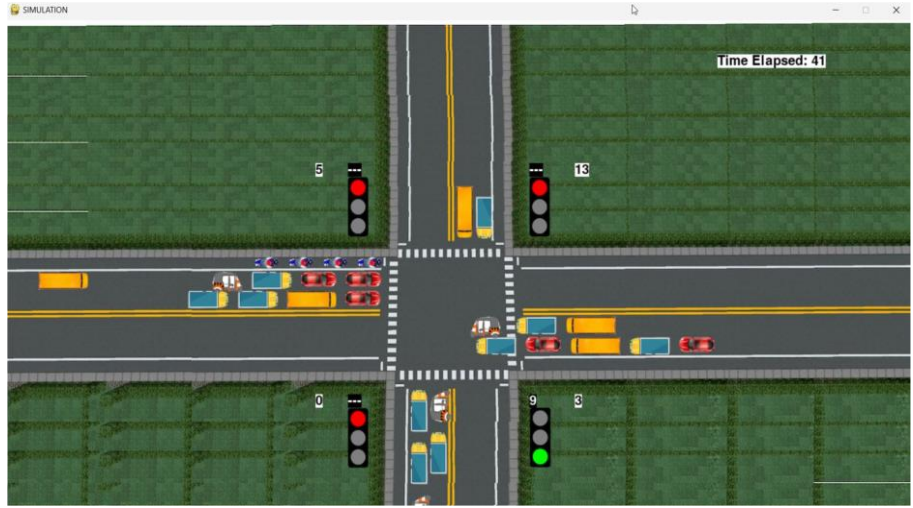

Figure. 4. Simulation Results.

The simulation displays the green signal time for vehicles moving upward to be 10 seconds, based on the vehicle count in that direction. This lane has significantly fewer vehicles compared to others. In a traditional static system, the green signal duration would be uniformly set to 30 seconds for all signals. However, this would result in inefficient use of time in this scenario. Our adaptive system recognizes the lower vehicle density and adjusts the green signal duration, accordingly, reducing it to 10 seconds to optimize traffic flow.

Comparing the traditional and proposed systems, we ran 15 simulations each for 5 minutes, the traditional system switches each traffic light every 30 seconds whether there are cars waiting to cross or not. Performance was recorded by counting the total number of vehicles crossing the intersection in the whole simulation.

The results in Fig. 5 show that the proposed adaptive system consistently out-performs the static system, no matter what the distribution. The level of improvement depends on how the probabilities vary. In simulations 2, 3, and 4, the probabilities of the car being in each lane are nearly equal, leading to slightly better performance compared to the static system, with an improvement of about 6%.

Moreover, when the traffic probability is evenly distributed among intersections, the proposed system outperforms the static system significantly. This improvement is evident in simulations 5, 6, 7, 8, 14, and 15, where performance increases by approximately 20%.

Finally, when the probability distribution varies widely between intersections, the proposed system shows a noticeable improvement compared to the current one. This can be clearly seen in simulation numbers 9 and 13, as shown in the graph in Fig. 5, where the blue line drops sharply. The performance is 34% better than the traditional system.

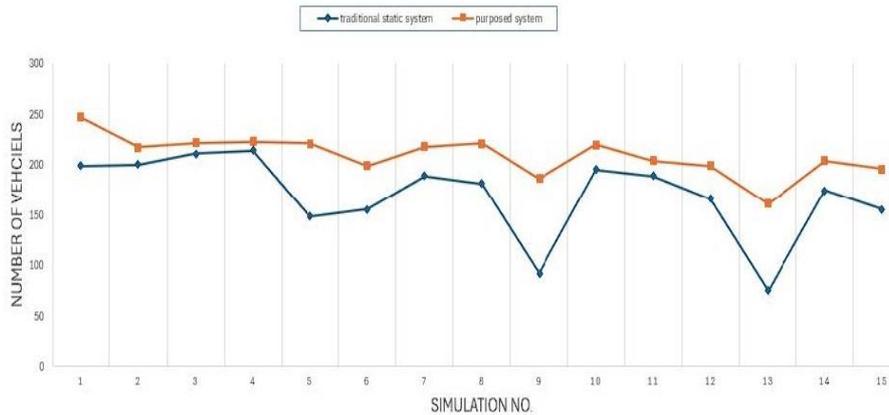

Figure. 5. Comparison between static and proposed system.

## 5. CONCLUSIONS

In this paper, a dynamic system for signal timings has been proposed. The pro-posed system adjusts traffic signal timings based on how much traffic is present at an intersection. This means the direction with more vehicles gets a green light for longer, helping to reduce delays, ease congestion, and shorten wait times. Which in turn lowers fuel consumption and pollution levels.

Simulation results show that this system improves intersection crossing efficiency by about 34% compared to traditional static systems. It also has several advantages over other advanced traffic control systems, like those using pressure mats or infrared sensors. A key benefit is its low cost since it works with existing CCTV cameras already installed at most busy intersections. In many cases, no extra hardware is needed beyond small adjustments to align the cameras.

Another advantage is the lower maintenance costs compared to pressure mats, which often break down from constant use. By integrating this system with CCTV cameras in major cities, traffic management can become more effective and efficient.

## AUTHORS

**Ahmed Mahmoud Elbasha**, is a fresh graduate from the electrical engineering department at the British University in Egypt. He has graduated in 2024 in the top 10% of his cohort. His research was focused on Vehicular communications and the use of AI in traffic Management.

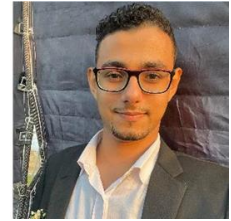

**Mohammad M. Abdellatif** (Senior Member, IEEE) is an Associate Professor at the Electrical Engineering Department at the Faculty of engineering, British University in Egypt (BUE). Dr. Mohammad holds a BSc '04 in Electronics and Communications Engineering from Assiut University, Assiut, Egypt, an MSc '06 in Telecommunication Engineering from King Fahd University of Petroleum and Minerals, Dhahran, Saudi Arabia, and a PhD '15 in Telecommunications engineering from the electrical and computer engineering department of the University of Porto, Portugal. His research interest includes Internet of Things (IoT), Cognitive Radio Networks (CRN) and Wireless Sensor Networks (WSN).

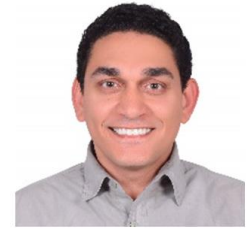